\documentclass[manuscript,screen,authorversion,nonacm]{acmart}
\AtBeginDocument{%
  }
\usepackage{marvosym}

\begin{document}

\title{The Cake that is Intelligence and Who Gets to Bake it: An AI Analogy and its Implications for Participation}

\author{Martin Mundt}
\affiliation{%
  \institution{University of Bremen \& Queer in AI}
  \city{Bremen}
  \country{Germany}
}
\thanks{Correspondence: Martin Mundt, University of Bremen, Germany, \Letter \, mundtm@uni-bremen.de}

\author{Anaelia Ovalle}
\affiliation{%
  \institution{Meta \& Queer in AI}
  \country{United States of America}
}

\author{Felix Friedrich}
\affiliation{%
  \institution{Technical University of Darmstadt \& hessian.AI}
  \country{Germany}}

\author{A Pranav}
\affiliation{%
  \institution{University of Hamburg \& Queer in AI}
  \country{Germany}
}

\author{Subarnaduti Paul}
\affiliation{%
 \institution{University of Bremen}
 \country{Germany}}

\author{Manuel Brack}
\affiliation{%
  \institution{Technical University of Darmstadt \& DFKI}
  \country{Germany}}

\author{Kristian Kersting}
\affiliation{%
  \institution{Technical University of Darmstadt \& hessian.AI \& DFKI}
  \country{Germany}}
  
\author{William Agnew}
\affiliation{%
  \institution{Carnegie Mellon University \& Queer in AI}
  \country{United States of America}}

\makeatletter
\let\@authorsaddresses\@empty
\makeatother

\renewcommand{\shortauthors}{Mundt et al.}

\begin{abstract}
\vspace{1em}  
\fontsize{9pt}{12pt}{\textbf{Abstract}} \newline
In a widely popular analogy by Turing Award Laureate Yann LeCun, machine intelligence has been compared to cake ---where unsupervised learning forms the base, supervised learning adds the icing, and reinforcement learning is the cherry on top.
We expand this ``cake that is intelligence'' analogy from a simple structural metaphor to the full life-cycle of AI systems, extending it to sourcing of ingredients (data), conception of recipes (instructions), the baking process (training), and the tasting and selling of the cake (evaluation and distribution). Leveraging our re-conceptualization, we describe each step's entailed social ramifications and how they are bounded by statistical assumptions within machine learning. Whereas these technical foundations and social impacts are deeply intertwined, they are often studied in isolation, creating barriers that restrict meaningful participation. Our re-conceptualization paves the way
to bridge this gap by mapping where technical foundations interact with social outcomes, highlighting opportunities for cross-disciplinary dialogue. Finally, we conclude with actionable recommendations at each stage of the metaphorical AI cake's life-cycle, empowering prospective AI practitioners, users, and researchers, with increased awareness and ability to engage in broader AI discourse.
\end{abstract}

\maketitle

\section{Introduction}
\label{intro}

\begin{quote}
	\emph{``If intelligence is a cake, the bulk of the cake is unsupervised learning, the icing on the cake is supervised learning, and the cherry on the cake is reinforcement learning.''} Yann LeCun, NeurIPS Conference 2016
\end{quote} 

\noindent By now an illustrious analogy in the field of machine learning (ML) and artificial intelligence (AI), the quote's original claim was that the bulk of the cake ---i.e., the majority of what supposedly composes intelligence--- does not rely on turning over exorbitant amounts of labeled data. On the contrary, the bulk is conjectured to consist of unsupervised learning (i.e., learning from data without labels), the icing is created through supervised learning (i.e., learning to match ground-truth annotations), and the ``cherry-on-top'' is reinforcement learning (i.e., tuning to feedback). 
In this cake metaphor, this resembles first baking a solid cake, if you so will the delightfully moist sponge, to lay the foundation for it to later be perfected to one's individual taste, i.e., a specific narrow AI task. If at first resulting in a skeptical smile, we will learn in this paper that the design of AI systems indeed has a lot of parallels to baking a cake. As such, we posit that the metaphor is valuable not for its original reasons, but rather because the creation of AI systems entails many of a real cake's associations and implications on ingredients, recipes, baking processes, taste, and consumers.   

In particular, we can follow the typical ML trends and may immediately ask ourselves whether this cake can be perfected? In turn, asking what the glazing on the cake would symbolize? In fact, perhaps it would even be possible to bake a cherry-chocolate cake from the start, reducing the number of steps along the way? Much of what is considered core AI advancement seems to revolve around these types of questions. Metaphorically speaking, it is refining the cake analogy to revise its recipe ---the algorithmic components to bake it---- and conversely contemplate its ingredients ---the data that goes into AI systems\footnote{See for instance the below medium blog post for an overview of how the AI cake analogy was first contested by \citet{Andrychowicz2017hindsightrep} to be composed of cherries (hindsight experience replay in reinforcement learning) and later revised to an AI cake 2.0 recipe by LeCun at ISSCC in 2019 by replacing fully unsupervised learning with the idea of self-supervision \url{https://medium.com/syncedreview/yann-lecun-cake-analogy-2-0-a361da560dae}}. From an ML perspective, baking the best possible cake may sound like an admirable goal. It certainly appears like a goal worthy of making progress if it were not for a substantial set of problems. For one, our goal implies measurability, the basis for it to later be optimized, despite being notoriously subjective in nature. As such, we can ask ourselves further questions, questions that are, however, of no less importance: \emph{``How do we acquire our ingredients?''}, \emph{``Who knows and understands the recipe?''} and \emph{``Who gets to bake the cake?''}. Thinking about these questions then creates a ripple effect to critically reflect on \emph{``Who decides if it is delicious?''}, \emph{``Who is allowed to indulge in it?''}, or \emph{``Who profits from its consumption?''}. Worst of all, if we do not like the current answer to any of these questions, is there anything we can do to change? For instance, what if we have personal reasons to object to the ingredients (say intolerance or ethical considerations) or do not like the final taste? Indeed, our proposition is not that the cake analogy is highly accurate because of its initial distinction between learning paradigms. Instead, we posit that it is valuable because once we ``stir ingredients and bake it'', there presently exist little tools to alter the final product. \\

Following the ``intelligence as a cake'' analogy, this paper critically questions how the metaphorical AI cake is presently being baked, shared, and eaten, where its ingredients originate from, and who is enabled to understand the recipe. To this end, we first re-conceptualize the analogy to describe how the process of translating baking ingredients to a cake relates to AI workflows. Having established accurate parallels, we create opportunities for cross-disciplinary dialogue by dissecting entailed issues and linking them to technical foundations. As such, we complement a rich history of works highlighting socio-ethical concerns by mapping where they are intertwined with fundamental technical challenges in AI design. Finally, we provide first actionable recommendations at each stage of the metaphorical AI cake's lifecycle, and in doing so, suggest avenues towards fostering participation and sustainability in AI design processes. 

\section{AI cake: an accurate analogy}
If the introduction left the reader hungry for more, they will hopefully see their hunger stilled in a transition to how the cake analogy translates to AI systems. To this end, let us first re-conceptualize the analogy to describe the various stages of the AI lifecycle and their social ramifications before proceeding to discuss underpinnings in technical limitations.

\subsection{Ingredients and their origin}
At the beginning of (baking) any cake are its ingredients. Depending on the type of cake and the baker's exact location, some of these ingredients may be locally available, i.e.,~acquired through trade or bought from a market, whereas others originate from far away. For instance, the use of cocoa seeds is prevalent in cakes of the northern hemisphere, yet its growth is restricted to equatorial (predominantly African) regions. Although the ``Western world'' now largely acknowledges that respective local cultures have been, and are still continuously, subject to exploitation, it generally remains challenging for the consumer to thoroughly understand the origin of ingredients and respective implications. There may exist nutrition labels, but they abstract away most information. More extensive pushes for transparency and ethical considerations, like the supply chain act \cite{SupplyChainAct2024} recently passed by the European Parliament, remain controversial among involved governments. 
At large, the pattern of obfuscation is exacerbated by ingredients having been processed before they reach the baker ---say, cocoa seeds turned into chocolate--- or even because the ingredients have been packaged into a ready-to-bake form in a chain of steps by big corporations. 

As much as cake relies on its ingredients, so do AI systems depend on their underlying data. Unfortunately, the analogy also extends from consumers remaining unaware of how ingredients have reached their plates, to AI system designers typically lacking full visibility into the sourcing and creation of training datasets. 
In both cases, companies (or an in-circle of stakeholders) embed their own biases into the gathering pipeline, dictating sourcing decisions that ultimately carry significant social ramifications.
The rapid expansion of machine learning datasets (e.g., \citet{Schuhmann2022Laion5b}) and increasingly opaque closed-source practices (e.g., OpenAI's ChatGPT) have only exacerbated these issues.
On the one hand, unchecked scaling of AI systems has thus led to increased exploitative practices. This is evidenced, for instance, by the outsourcing of psychologically harmful content to African workers for moderation and annotation \cite{Rowe2023MentalTollAIModels}, a malpractice labeled as ``ethics dumping'' by the EU Horizon Research Council \cite{EuropeanComission2015EthicsDumping,Schroeder2018EthicsDumping}.
Similarly, ChatGPT's ``surprisingly'' frequent use of terms like ``delve'' was later attributed to the terminology's prevalence in Nigeria \cite{Hern2024CheapOutsourcedAILabor}. 
On the other hand, the lack of transparency and access to the data pipeline has severe implications for data ownership or copyright infringement. As an example, the work ``Does CLIP know my face'' \cite{Hintersdorf22DoesClipKnowMyFace} has raised attention to the fact that people's personal data may indeed have seeped into the metaphorical AI cake as an ingredient without consent, but it remains very challenging to find out about it. 
Consequently, such model applications have the potential to violate privacy laws by retaining personally identifiable information, resulting in the first legal actions against the corporations behind them \cite{Noyb2024ChatGPTFalseInfo}. 

Clearly, both real cake and its AI analogy thus hinge on their ingredients. Where colonialism has facilitated the exploitation of natural ingredients, amassing data seems to follow a reminiscent pattern based on digital imperialism.
As the origins of AI’s ``ingredients'' remain obscure and processes remain opaque, the lack of transparency accelerates the privatization of knowledge \cite{Ferrari2023FoundationModKnowledgePrivatization}. The underlying pattern of abstracting away the contextual complexity and ecosystem removes a sense of responsibility ---understanding that humans are behind--- as it gets easier in practice to exploit because we can neither see nor retrace. 

\begin{figure}
    \centering
    \includegraphics[width=0.95\linewidth]{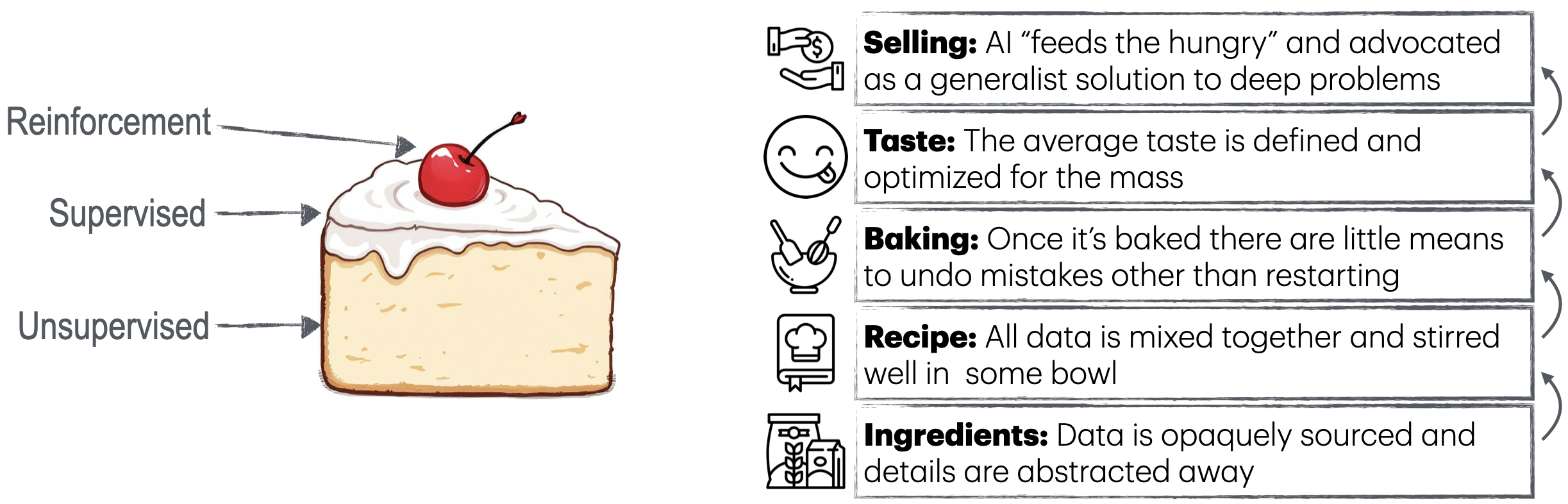}
    \caption{An illustration of the AI--cake analogy. (Left) Traditionally, the cake was used to provide a structural metaphor for machine intelligence, relating unsupervised (bulk), supervised (icing), and reinforcement learning (cherry) paradigms. (Right) Our re-conceptualized analogy extends the original metaphor by drawing parallels to the way AI's ingredients are sourced, recipes crafted, and ultimately how the metaphorical cake is baked, tasted and sold.}
    \label{fig:teaser}
\end{figure}

\subsection{The ``best'' cake recipes}
Baking a cake does not only require a set of ingredients, it generally needs to follow a recipe. The latter typically ensures that chosen ingredients ``interact'' sufficiently well. Although there exists an abundance of instructions on how to combine a cake's ingredients, the individual ingredients eventually all fall victim to the same blending process when mixed together into dough. Remarkably, this process seems simultaneously accessible and demanding. 
As long as the recipe is followed closely, almost anyone can successfully bake a cake, provided they have the massive ovens to bake it (see baking process section). 
If, however, one wishes to include a particular new ingredient in the mix, then baking almost becomes a precision science.  It is a delicate balance of the modifications being noticeable, e.g., in color, texture, or taste of the cake, or potentially overpowering and thus ruining the mix ---a balance only few may claim to truly understand. It certainly becomes a futile effort if one wishes to turn cake ingredients into another dessert when only generic cake recipes are available.

As much as cake recipes ultimately combine the majority of ingredients together, so do AI systems mix their training data. Unfortunately, our analogy extends from the blending procedure relying on the predominant belief that adding more data to the metaphorical dough is sufficient for inclusion, to the fact that few experts, if any, understand the consequences of respective interleaving. In both parts of the analogy, the assumption that simply adding more data ensures inclusivity is flawed. 
\citet{Delgado2021diverse_stir} have pointed out that \emph{``we cannot just add diverse end-users and stakeholders and stir''}. 
On the one hand, it is unclear which combinations of data may advance model capabilities and which data additions entail social consequences; their nature hinges on complex interplay with the rest of the opaque data mixture. For instance, from the perspective of purely instrumental improvement, it is heavily debated when and how the addition of synthetically generated data is indeed beneficial, e.g. Phi-2 \& 3 \cite{Javaheripi2023Phi2,Abdin2024Phi3} or Dall-e-3 \cite{Betker2023Dalle3}, or whether its inclusion in the mix is useless or downright adverse to performance \cite{Hao2024Syntheticdataaichallenges}. From a complementary normative angle, initially sincere efforts to ``debias'' models through increased addition of non-anglocentric languages have similarly concluded that stirring these ingredients has contributed fairly little towards the desired goal of broad applicability in globally inclusive perspectives \cite{Friedrich2024multilingualtexttoimagegenerationmagnifies}. On the other hand, the recent repercussions \cite{Raghavan2024GeminiGotItWrong} surrounding Google DeepMinds' Gemini image generation \cite{GoogleGemini2024} have rendered the challenge of highlighting ingredients in a recipe and unwillingly overshooting perfectly visible\footnote{and further lacking sufficient input of how people would in principle like to be depicted - explored in-depth in the ``what makes for a tasty cake'' section}. In short, an effort to ``diversify'' the generated images through naive integration attempts of broader coverage of society has further resulted in the formation of adverse relations \cite{Robertson2024GeminiNaziGen, Gilliard2024GeminiDeeperProblem}, e.g. now picturing strong racial diversity in the generation of Nazi Germany's Wehrmacht soldiers. This contrived effect is not only highly illogical but further contributes to perpetuating harmful AI content and fostering oppression in complete contradiction to the initial amiable aim \cite{Noble2018AlgorithmsOfOppression}.

Clearly, both real cake and its AI analogy thus hinge on combining their ingredients. Whereas careful measuring and stirring of ingredients is a challenging task already, complex interplay of data makes careful weighting only one imperative element of a recipe. A lack of understanding in interleaving arbitrary data in AI systems yields mixtures where inclusion of ingredients, even if added with the best of intentions, can range anywhere from completely inconsequential, to instrumentally beneficial, or resulting in adverse fallout. The seemingly accessible and effortless aggregation of all data to streamline the ensuing baking thus also obfuscates ingredients' uniqueness, even suppresses their critical nuances, and in consequence gives rise to counter-intuitive ramifications. As such, homogenization of the recipe, i.e. blending all ingredients together to follow the instructions for cake dough (training a deep machine learning model), entails accessibility advantages that simultaneously make it excessively hard to bake anything else.    

\subsection{Cake foundation and the baking process}
Once settled on ingredients and a recipe, baking a cake is very much a unidirectional process. Once in the oven, there is no turning back. Forgotten an ingredient, bake a new one. Took a slight misstep in following instructions, bake a new one. 
Do not like the outcome, bake a new one.
Once baked, a cake's composition becomes fixed and permits few revisions.
Incorporating any raw ingredients post-baking is impractical and far from appetizing --- think for instance of pouring milk over or adding raw egg to a baked cake.
At best, we can superficially tune the cake, e.g. adorning it with fruit, candy, or a touch of glazing to make it more appealing.

As hard as it is to change a baked cake, so is modifying an AI model after it has been trained.
Within the preference alignment literature, the "superficial alignment hypothesis" posits that reinforcement learning from human feedback primarily affects a foundation model's \cite{Radford2019GPT2,Touvron2023llama2} textual output style, rather than its core capabilities \cite{Zhou2023SuperficialAlignment}.
In fact, respectively finetuned models result in near identical performance of tuned and base model versions (i.e., they trade off one dimension for another), at best resulting in superficial improvement \cite{Lin2024RethinkingLLMAlignment}. 
This is analogous to how we can only decorate a finished cake rather than change its composition, resulting in an unreasonable amount of baking repetitions for every fundamental change. In this context, however, such modification attempts (i.e. re-training) cost millions of dollars \cite{Koetsier2023ChatGPTburnsmillions,Vanian2023ChatGPTCosts} and consume extraordinary amounts of energy for minor improvements \cite{Strubell2019energy,So2019EvolvedTransformer}\footnote{We note that the baking of the metaphorical AI cake is not only unsustainable because of the baking process itself, but is also subject to sustainability considerations regarding the equipment required for baking, similar to our sustainability concerns regarding data ingredients. The scarcity of essential hardware, particularly GPUs, concentrates power among a few actors. Furthermore, manufacturing these components requires rare minerals, 
often sourced from conflict regions, adding another layer of sustainability concerns to AI development.}.
For example, while presumably being updated to improve certain model abilities in Spring 2023, ChatGPT actually lost proficiency in basic tasks like identifying prime numbers or writing simple code \cite{chen2023chatgptforgets} by June of the same year. This reveals a fundamental tension with current regulatory frameworks, such as the Biden-Harris AI executive order \cite{BidenHarrisAI,BidenHarrisAIIEEE} and the United Kingdom's ``pro-innovation'' approach \cite{UKProInnovationAI}, which rely heavily on post-training interventions.
While model auditing and red-teaming are crucial to identifying problems in AI systems, resolving them thus remains challenging due to the difficulty of modifying a model after training --- much like trying to fix a single ingredient in an already baked cake. These effects are further exacerbated by the fact that, following the prior section's arguments, we lack understanding of which data corresponds to the metaphorical egg or milk, and which ones are decorative in nature.

Overall, both real cake and its AI analogy thus hinge on a single-cycle baking process. Whereas baking cannot be fully changed and undone once completed, AI training largely dictates the final utility of the system. Attempts at fine-tuning are either superficial or entail strong trade-offs. In particular, later attempts at adding entirely new ingredients can interfere catastrophically, depending on the nature of the ingredient. Contrary to intelligence also being described as \emph{``the ability to adapt to change''} (commonly attributed to Stephen Hawking, see \citet{Strauss2018hawking}), the current inflexibility of the ``cake-like'' training pipeline entails excessive process repetition. As each training run requires exorbitant amounts of computational resources \cite{Ludvigsen2023GPT4CO2,Patterson2021LargeScaleCarbon}, this seems akin to baking cakes by having thousands of ovens emit carbon, to ultimately re-do the entirety of produced cakes any time a non-superficial adjustment needs to be added.

\subsection{What makes for a tasty cake?}
A cake may have been baked successfully, but is it also tasty? Some flavors may widely be assumed to be ``safe bets'', like chocolate seemingly enjoying popularity, but who gets to provide this assessment? 
Imperial history may have imposed particular cuisine aspirations upon us, but what is realistically considered delicious will vary drastically. When considering geographical and cultural influences, it will inevitably be impossible to single out one taste.
Resorting to stereotypical hyperbole for clarity, strong sweetness may for instance be desirable in portions of the world, but much less strongly desired in other parts. Certainly, the notion of an average delightful cake itself is a poor simplification even within one region. 
After all, every human has their own flavor preferences; preferences that are uniquely shaped by their upbringing and further influenced by constraints such as food intolerances or allergies. And in the end, different parts of the world may also prefer entirely different deserts or be more cautious of a potentially unhealthy diet.

As much as it is impossible to define a best-tasting cake, so is it impossible to define the best-performing AI.
Grappling with how to integrate these preferences is a recurring exercise throughout machine learning research, as demonstrated by the subjective nature of human feedback in learning preferences \citep{kirk2024prism}.
For instance, in large language modeling, a model's performance could typically be determined through a set of prescriptive benchmarks, which predominantly center common-sense reasoning \cite{zellers2019hellaswag, ai2:winogrande, talmor-etal-2019-commonsenseqa}, reading comprehension \cite{rajpurkar-etal-2016-squad, clark2019boolq, choi-etal-2018-quac}, or mathematical abilities \cite{cobbe2021gsm8k,lightman2023lets}. 
While these evaluations provide valuable insights into the capabilities of AI models, they fail to capture the full complexity of human preferences and the socio-technical implications of deploying these models in real-world contexts, including the impact on historically targeted, marginalized, or economically underprivileged groups. As such, gender bias literature is typically restricted to a binary lens, limiting critical discourse on AI harms deriving from a focus on binary gender \cite{Queerinai2023, ovalle2024root} and primarily male perspectives in AI systems \cite{ovalle2023m, dev2021harms}.
Likewise, the famous GenderShades audit \cite{Buolamwini2018GenderShades} revealed that commercial AI systems often misclassified darker-skinned females \cite{Birhane2022UnseenBlackFaces}. A recent WIRED article \cite{Wired2024SoraQueer} further illustrates the collapse to an expected average, showing how OpenAI's video generator Sora \cite{OpenAI2024Sora} defaults to depicting bisexual, asexual, or transgender persons with pink hair --- highlighting the need for increased demographic representation and intersectional measurement practices in the baking process. These challenges are compounded by the influence of those who decide which AI technologies are put into practice and how acceptable performance is defined --- i.e., what the baker determines as the average taste gives rise to power \cite{Bommasani2022EvaluationChange}. On the other hand, expanding a baker's palette to make a tastier ``AI'' cake often presents a challenging and sometimes contradictory undertaking, as evidenced by the more than 21 definitions of fairness \cite{Narayanan2018Fairness21Def}. Achieving the latter simultaneously proves impossible, especially when group and individual fairness conflict \cite{Kleinberg2017FairnessImposibility, Binns2020FairnessConflict}. 
At the same time, the social implications of each are unclear, as each method brings its own unique set of fairness challenges \cite{Castelnovo2022FairenssNuances, Lum2022DebiasingBiasMeasurement}. 
In turn, the research community has proposed over 70 AI fairness evaluation guidelines \cite{Bellamy2019AIFairness360}.

Overall, both real cake and its AI analogy thus hinge on subjective and even incompatible notions of being delightful. Just as a bakery attempts to cater to the average customer's preferences, AI systems often rely on a limited understanding of typicality or correctness, leading to a collapse of diversity into oversimplified norms. Faced with complexity and abundance of taste assessment, AI bakers may opt for measures that are easy to satisfy or superficially appealing. This facilitates "ethics shopping" or "ethics bluewashing" \cite{Floridi2019FiveRisksUnethical}, respectively cherry-picking what can be satisfied and using superficial measures in favor of positive appearance.

\subsection{``If they do not have bread, let them eat cake!'' --- Sharing and (over-)selling}
The reader may finally chuckle when reading about the metaphor's last connection to the above infamous quote, attributed to Marie Antoinette (and likely stated by Jean-Jacques Rousseau in 1765). But like any product, cake is eventually shared. More likely, it will be sold in order to make up for the initial cost and earn enough profit to make a living. Initially, this may have a naive benevolent intention, much like in our 18th-century anecdote, but there certainly is little use to suggest a starving population bake cake. After all, as elaborated in the prior sections, they will lack access to ingredients, lack understanding of the recipe, lack the tools to bake, or derive little nutritional value from the cake.

As much as the starving population of the 18th century had more basic needs than cake, so does much of the current population not actually profit from claimed AI progress. On the one hand, this may partially be due to a severe mismatch between the generally sold capability, and what is practically specified in the AI creation process. For instance, ``diverse representation'' is frequently advocated for greater inclusion of underrepresented groups in datasets, yet oversimplified notions lead to objectification or exploitation \cite{Bergman2023RepresentationAIEval,Chien2024RepresentationalHarms}. Similarly, common error rate measures often lead to misleading promises on fair systems, given that assessment does not account for ultimately entailed effects \cite{Corbettdavies2023MismeasureFairness}. This leads to treatment and impact disparity, where correlations arise unintentionally and outcomes start to differ across groups \cite{Lipton2019MLDisparity}. On the  other hand, falsely promised profit may also be due to a growing belief in technosolutionism: the belief that technology is always the solution \cite{Broussard2019ArtificialUnintelligence}. In this belief, an algorithm's role in selective targeting may simply be neglected at the prospect of later improvement. As such, periodic examples of how tech exacerbates inequality can be found. In fact, it generally seems that many AI contributions that were sold with some initial notion of good in mind, ultimately ended up fostering undesired population surveillance \cite{Kalluri2023SurveillanceAIPipeline}, all the way to the latter being used to AI-enabled direct persecution \cite{Harwell2020Uighur}. 
Sometimes the technosolutionist narrative's harm may be invisible on the surface, especially in scenarios where the AI cake is oversold to users that presently don't require it at all. An intuitive example of this pattern is the frequent marketing of AI as an opportunity to establish food security and enhance health care, in particular on the African continent \cite{Arakpogun2021AIAfricaOppChall}. Even if the potential benefits of AI are enormous, it is also challenging to reap them while ignoring the structural inequalities that need to be overcome before AI can actually live up to its sold promise \cite{Ade-Ibijola2023AIAfricaChallenges}. Dynamic power relations between countries are not easily resolved by AI and general power imbalance cannot be overlooked. For instance, the US and EU heavily subsidize agriculture to export to African consumers \cite{Komminoth2023AIAfricaAgriculture} and big tech corporations own an overwhelming amount of infrastructure. Even when local grassroots progress is thus made, external parties and companies end up reaping many of its benefits - a pattern referred to as cooptation \cite{Mwema2024UnderseaAfrica}.  

Overall, both distributing cake to a starving population and overselling AI systems hinge on an initially amicable incentive that fails to deliver on its intended goal. Whereas suggesting to feed people with cake is a poor nutritional and mismatched solution, so too are AI systems frequently oversold beyond challenges they are capable of solving. Technosolutionism and the resulting practice to create terminology to fuel this belief, e.g., ``foundation models'' or ``frontier AI'' \cite{Helfrich2024FrontierAIHarms}, shift away focus from actual current capabilities to long-term speculation. In turn, a habit of ``ethics shirking'' \cite{Floridi2019FiveRisksUnethical,Nasdaq2024Shirking} is facilitated, where less work on ethical aspects is conducted if less hypothetical reward is expected. 

\section{Technological underpinnings: Why it is difficult to change the AI cake}
The previous sections have substantiated the AI cake analogy and have linked it to several concerning practices. We now revisit these ramifications and map where social outcomes are intertwined with and exacerbated by technical underpinnings. In particular, we posit that even if consensus on instrumental and normative angles to AI development existed, the present technical foundations make it tremendously challenging to translate benevolent aims into practice. 

\begin{figure}
    \centering
    \includegraphics[width=0.925\textwidth]{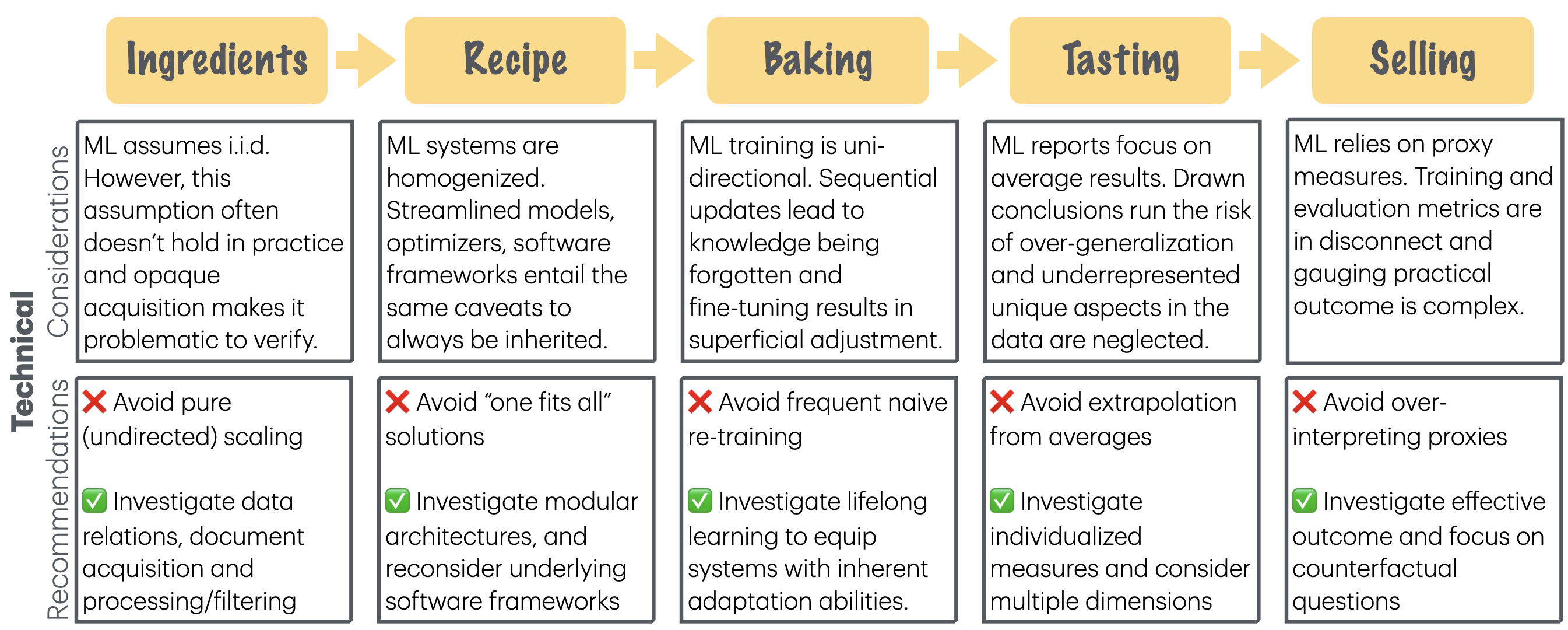}
    \caption{An overview of section three's  considerations with respect to the AI-cake's technical foundation and our future recommendations for each process stage from ingredients and recipes, to baking, tasting and selling. 
    }
    \label{fig:recommendations}
\end{figure}

This is not to say that researchers should be exempt from any responsibility. On the contrary, we precisely wish to empower them with a thorough understanding of how their present technical choices imbue constraints, foster biases, and why fundamental (mathematical) properties at the heart of (statistical) machine learning imply significant barriers. Mirroring our earlier paper structure, we begin each subsection with a quote from a recent publication to exemplify one key technical hurdle underpinning the previously described challenges, before proceeding to disentangle its technical caveats. We respectively finalize each subsection with a set of technical recommendations for future work, to highlight cross-disciplinary opportunities alongside existing social and ethical research advances. Figure \ref{fig:recommendations} provides a summary.

\subsection{``Ingredients'': the challenge of data ``i.i.d.-ness''}
\begin{quote}
\emph{``Our results show that hate content increased by nearly 12\% with dataset scale, measured both qualitatively and quantitatively using a metric that we term as Hate Content Rate.... This, as we hypothesize, may be a consequence of rich non-i.i.d. inter-sample correlations emerging from a graph-structured prior for CommonCrawl''} - \citet{Birhane2023laions}
\end{quote}

\noindent The above quote refers to the scaling of data examples from the LAION-400m \cite{Schuhmann2021Laion400m} to the LAION-5b \cite{Schuhmann2022Laion5b} dataset (respectively containing 400 million and 5 billion data points), attributing increasingly hateful content to the lack of understanding of non-i.i.d. correlations in the data selection and filtering mechanisms. 
More formally, i.i.d. refers to ``independent and identically distributed''. The U.S. National Institute of Standards and Technology (NIST) provides a concise definition: ``A quality of a sequence of random variables for which each element of the sequence has the same probability distribution as the other values, and all values are mutually independent'' \cite{NIST2018IID}. In other words, we assume each data point to be different from and unaffected by others, while all data points are expected to originate from a common data generation process. Intuitively, this further entails ``exchangeability'', i.e., the notion that we can exchange the order of our data points in practice ---  a property that will also be central to our later learning recipes.

Naturally, there are various ways in which the i.i.d. assumption won't hold in the real world. Unfortunately, violations will mostly occur in unknown ways, with each respectively obscuring our understanding of the gathered data. 
An easy example is if the acquired dataset selectively contains (near) duplicates. This has recently occurred for up to 30\% of LAION-2b \cite{Webster2023deduplication}, but can be dealt with effectively through various statistical tests \cite{Hutter2022testingindependenceexchangeablerandom}. It becomes significantly more challenging when a) there are complex inter-dependencies between subsequent data points, b) the distribution changes over time and becomes non-stationary, c) the data is adversarially crafted \cite{darrell2024machine}. 
Although the latter scenarios are well-known to be realistic, the i.i.d. assumption is rooted deeply in our algorithms for pattern recognition and in statistical learning theory \cite{vapnik199learningtheory}. It is a key requirement to render many algorithms practical, in particular in the prevalent machine learning angle to AI. Specifically, i.i.d.-ness of data is presumed because it provides a required simplification of the likelihood function, the essential tool underlying learning/optimization procedures:
\begin{equation}
    \mathcal{L}(\mathbf{\theta}) = p(\mathbf{x_1},\mathbf{x_2},\ldots,\mathbf{x_N}|\mathbf{\theta}) = p(\mathbf{x_1}|\mathbf{\theta})p(\mathbf{x_2}|\mathbf{\theta}) \ldots p(\mathbf{x_N}|\mathbf{\theta}) 
\end{equation}
Here, the i.i.d. assumption is critical because it allows to convert an initially complicated to calculate function, where we wish to infer parameters $\mathbf{\theta}$ that describe a full set of data points $\mathbf{x}_n$, into a product of terms based on respective individual data points. In fact, if we move to a logarithmic space, the product even becomes a simple sum:
\begin{equation}\label{eq:empirical_loss}
    \log \mathcal{L}(\mathbf{\theta}) =\log p(\mathbf{x_1}|\mathbf{\theta}) + \ldots + \log p(\mathbf{x_N}|\mathbf{\theta})
\end{equation}
In turn, computation becomes manageable at the cost of neglecting data inter-dependencies. In fact, even in scenarios where such inter-dependencies are directly obvious, for instance, in the naturally temporally ordered data streams of reinforcement learning, a host of tricks are introduced to leverage i.i.d.-ness. For the particular example, it is typically experience replay that is used, where a subset of old observations is stored in a memory buffer that then resembles a typical i.i.d. dataset \cite{Rolnick2019ExperienceREplay}. As such, the oversimplification of datasets and the equal mixture of all data ingredients is a direct result of the steps seemingly necessary to render computation feasible. The lack of technical tools to discover independence and discover causal relations at scale \cite{Scholkopfetal21}, coupled with the fact that real-world datasets are frequently inaccessible or in-transparently acquired, exacerbate the opaqueness. \\

\noindent \textbf{Recommendations ---} \textit{towards understanding and tracing of ingredients:} \\
A central recommendation is to avoid over-reliance on the ``unreasonable effectiveness of data set scale'' \cite{Sun2017unreasonabledata}. As evident from the critical history of, e.g., ImageNet \cite{Denton2021GenealogyDatasets,Deng2009ImageNet},  datasets require change over time. Whereas outliers in the form of mislabeled examples may pass as i.i.d. by reducing their prediction variance through large training sets, systematic correlation will reciprocate systemic bias at any scale \cite{dundar2007learning}. It is, thus, essential to approach datasets more carefully than random scraping and to meticulously document the processes surrounding its acquisition, e.g., using resources such as data statements \cite{Bender2018DataStatements,Gebru2018Datasheets}.  
If the i.i.d. assumption is computationally necessary, then additional processing - such as clustering and removal - should be performed to warrant the assumption. These early initiatives should be extended to encompass detailed analysis of non-i.i.d. relations and enable digital data tracing. This will improve understanding of included data in a system, both in terms of technical assumptions and to expose malicious practices, such as stealing.

\subsection{``Recipes'': Homogenization of instructions } 

\begin{quote}
    \emph{``Though foundation models are based on standard deep learning and transfer learning, their scale results in new emergent capabilities, and their effectiveness across so many tasks incentivizes homogenization. Homogenization provides powerful leverage but demands caution, as the defects of the foundation model are inherited by all the adapted models downstream.''} - \citet{Bommasani2021FoundationModels}
\end{quote}

\noindent The above quote emphasizes the increasing homogenization as a result of deep learning's success at scale, but also advises caution with respect to always inheriting the same caveats. 
The latter is primarily driven by the seeming possibility to employ a heavily standardized framework of deep models, nowadays transformer \cite{Vaswani2017AttentionAllYouNeed} based foundation models, and optimization through backpropagation \cite{Werbos1982Backprop,Rumelhart1986Backprop,LeCun2012EffBackprop}. As deep neural networks are known to be universal approximators \cite{HORNIK1989359}, this fosters a narrative of ``one model to learn them all'' \cite{Kaiser2017ModelLearnAll,Reed2022Gato}. Although homogenization offers several benefits in terms of accessibility and ease of use, it suffers from oversimplifying and locking in the recipe. 

From a technical perspective, it is a result of modeling any modern neural network as a cascade of layers $l = 1, \ldots, L$ that map from an initial dimension $d_{l-1} \in \mathbb{N}_{l-1}$ to a resulting arbitrary dimension $d_{l} \in \mathbb{N}_{l}$ through a progressive set of transformations $T_l: \mathbb{R}^{d_{l-1}} \rightarrow \mathbb{R}^{d_l}$ to assemble a ``hypothesis'' (model) $h_{\theta}$:
\begin{equation}
    h_{\theta}(\mathbf{x}) = T_L(T_{L-1}(\ldots(T_1(\mathbf{x})))), \quad x \in \mathbb{R}^{d}
\end{equation}
This hypothesis is then learned to match the data evidence with a common set of optimization algorithms belonging to the (stochastic) gradient descent family, such as the tremendously popular Adam optimizer \cite{Kingma2017Adam}. Concerns over the limitations, for instance, the fear of overfitting to the data, i.e., the technical equivalent of ``learning by heart'', seem to be surmounted by homogenization at scale through the double descent phenomenon \cite{Nakkiran2020DoubleDescent}\footnote{The observation that training with substantial data amounts for prolonged periods of time eventually overcomes any initial performance deterioration}.
In addition to the social implications of the homogenized recipe ---recall ``we cannot just add diverse end-users and stakeholders and stir'' \cite{Delgado2021diverse_stir}--- the entailed technical limitations of this homogenization can best be understood in its manifestation in  programming frameworks. Over time, the original differences of pioneering machine learning software frameworks, such as Theano \cite{TheTheanoDevelopmentTeam2016} and Torch7 \cite{Collobert2011Torch7}, have converged in terms of functionality. As such, whereas there may be syntactical differences and functionality nuances, current age software ---e.g. the prevalent Pytorch \cite{Paszke2019PyTorch}, Tensorflow \cite{Abadi2016TensorFlow}, and Jax \cite{Jax2018Github}--- is severely streamlined towards deep neural networks. This makes breaking out of homogenized recipes more than just challenging, as noted recently by a work with the provocative thesis ``ML is stuck in a rut'' \cite{Barham2019MLStuckInARut}. In this work, they show that an attempt at more modular, more brain-like architectures, for instance, the seminal Capsule network \cite{Sabour2017Capsules}, underperformed the homogenized recipes. Alas, the catch is that this is not necessarily due to inferior design, but rather a direct function of programming frameworks being excessively tailored to homogenized deep learning recipes. Despite employing very similar mathematical functions (i.e. convolutions), Capsule networks thus appeared significantly worse than they practically should be. This tight coupling between a narrative that a homogenized recipe is ``all we need'' \cite{Vaswani2017AttentionAllYouNeed,Bommasani2021FoundationModels} with focused tailoring of software frameworks exacerbates the challenge of creating any other solutions. As such, overcoming the challenges of our homogenized recipe is not only conceptually challenging but also exacerbated on a technical level through excessive convergence of underlying tools. \\

\noindent \textbf{Recommendations ---}\textit{towards unique and customizable recipes:}\\
A central recommendation is to avoid over-reliance on homogenized foundation models with streamlined software. Although there is potential for rapid applications, the underlying convergence disincentivizes the development of breakthroughs in understanding or effective small-scale solutions. As such, we recommend research on modular layering and inspectable representations, for instance by furthering theoretical understanding of information flow \cite{SchwartzZiv2017Information,Saxe2018InformationBottleneck} instead of hoping for ``emergence'' of capabilities. For a more radical recommendation, we recommend questioning whether a homogenized model is suited to individual needs and considering whether auxiliary non-data-driven methods may be advantageous. Any such alternative approaches are not easy to implement in existing software frameworks, similar to our earlier Capsule reference. To give but one example, we can draw inspiration from neurogenesis in mammalian brains \cite{Gross2000Neurogenesis,Vadodaria2014Neurogenesis} - shrinking and growing neurons on the fly - and translate this ability to dynamically adapt neural structures to the task at hand \cite{Ash1989DynamicNode, Evci2022GradMax, Mitchell2024SelfExpanding}, requiring diversified development of programming frameworks.

\subsection{``Baking process'': the interference dilemma}
\begin{quote}
    \emph{''Human learning has evolved to thrive in dynamic learning settings. However, this robustness is in stark contrast to the most powerful modern machine learning methods, which perform well only when presented with data that are carefully shuffled, balanced, and homogenized.``} - \citet{Hadsell2020EmbracingChange}
\end{quote}

\noindent The above quote emphasizes how fluid intelligence allows humans to excel in dynamic environments through sequential adaptation and progressive refinement of their knowledge \cite{Flesch2018continualmindmachine,Flesch2023ContinualContextGating}. In contrast, the strength of (large) machine learning models is only apparent through crystallization of carefully balanced and homogenized data. Whereas it is significantly easier for humans to continually learn ``the $n$-th thing'' after learning ``$n-1$ things'' \cite{Thrun1996LearningNth,Hadsell2020EmbracingChange}, human-like knowledge transfer in machines remains a challenging desideratum \cite{Pan2010Transfer, kudithipudi2022biolifelong}.

From a technical perspective, the lack of ability to learn continually entails an exorbitant amount of retraining. Yet, this re-training cost is willingly embraced, as the unfortunate alternative is induced catastrophic forgetting \cite{McCloskey1989Catastrophic,Ratcliff1990}. The latter refers to the phenomenon of abruptly losing acquired information when sequentially tuning to new data. Unfortunately, the reasons for this phenomenon are deeply ingrained in our optimization toolkit, where one culprit is the iterative nature of our optimizers \cite{Tsypkin1966Training}. At the example of our homogenized recipe's stochastic gradient descent, steps $\tau$ of updates to parameters $\theta$ are conducted in a loss function across observed data points with a ``learning rate'' $\eta$: 
\begin{equation}
    \mathtt{for} \, \, \tau = 1,2, \ldots, t \, \, \mathtt{do:} \quad
     \mathcal{\theta} \leftarrow \mathbf{\theta} - \eta \nabla \mathcal{L}_{\tau}(\mathbf{\theta})
\end{equation}
Intuitively, this works well if we present the optimizer with all the concepts we ultimately care about. The optimizer will ``move in directions'' that satisfy all observed concepts and progressively improve. However, it is now similarly unsurprising that if presented with a novel fraction of data at a later time step, this optimizer will move uni-directionally to improve on only what it has recently seen. This challenge is significantly exacerbated by the semi-distributed nature of representations in neural models \cite{French1992SemiDistributedForgetting}. Although entangled representations foster generalization across concepts by moving away from a look-up table, they also imply that most any update has an influence on every learned concept so far.  
The field of continual machine learning \cite{Hadsell2020EmbracingChange,Mundt2023WholisticCL} aims to overcome this central limitation. Alas, it is presently bounded by our homogenized model and optimizer recipes. If we cannot leave the frame, the metaphorical cake can only be changed superficially post-hoc or re-baked fully when we wish to make major additions or changes. Learning becomes a (Markov) chain that only takes into account the most recent past. It does not explicitly take into account all prior observations, making any addition either superficial, or potentially catastrophic, if the desired change was not part of the original data mix. \\

\noindent \textbf{Recommendations ---} \textit{towards adaptive and collaborative learning:} \\
A central recommendation is to avoid unsustainable re-training advice, e.g. Google Cloud's \cite{GoogleCloud2021Workflow} ``Developing a model is a process of experimentation and incremental adjustment. You should expect to spend a lot of time refining and modifying your model to get the best results.'' Although deep learning representations have historically been argued to rival primate IT Cortex (for vision) \cite{Cadieu2014DLCortex}, their ability to ``embrace change'' \cite{Hadsell2020EmbracingChange} certainly falls far behind that of any human \cite{Flesch2018continualmindmachine, Chen2018Lifelong}. In turn, we recommend drawing inspiration from lifelong learning mechanisms to equip AI with efficient adaptation capabilities, absolving us from high re-training costs. On the one hand, this requires translating the plethora of biological capabilities known to contribute to lifelong learning to artificial systems \cite{kudithipudi2022biolifelong} and establishing respective AI theory \cite{Prado2022ContinualTheory}. On the other hand, a revisit to neuro-symbolic systems, that in parts were able to adapt sustainably throughout their entire life-cycles by interfacing learning with reasoning \cite{Carlson2010NELL,Chen2013NEIL,Mitchell2015neverending}, may be warranted.

\subsection{``Taste'': limits of averages and aggregates}

\begin{quote}
    \emph{``Some pitfalls are only visible when examining the dataset as a whole or the proposed aggregating metrics. Since what the benchmarks aim to measure is not well articulated, it can be difficult to distinguish whether and when the pitfalls we list below suggest a poor conceptualization of stereotyping or instead call into question the way it is operationalized''} - \citet{Blodgett2021StereotypingSalmon}
\end{quote}

\noindent The above quote, originally written in the context of the misattribution of stereotypes, points to issues in the interpretation of averaged assessment. It implies that an AI system's pitfalls may become obfuscated, for instance, an underrepresented group not being adequately covered, and that important nuances may typically be neglected, such as through aggregating fairness metrics \cite{Castelnovo2022FairenssNuances}. The present over-reliance on aggregate measures \cite{Burnell2023RethinkReporting} respectively makes it challenging to predict practical behavior in real situations outside benchmark datasets.

From a technical perspective, the popularity of average assessment is not only driven by the scientific literature's seeming necessity to compare single benchmark numbers, but the challenges are once more rooted deeply in the algorithmic underpinnings themselves. Following earlier equation \ref{eq:empirical_loss}, which highlighted the significance and prevalence of the i.i.d. assumption, evaluation losses ($\mathcal{L}$) and measures are  averaged in equal weighting of all data points $n = 1, \ldots, N$: 
\begin{equation}
    \mathcal{L}(\mathbf{\theta}) = \frac{1}{N} \sum\nolimits_{n=1}^{N} \mathcal{L}_{n}(\mathbf{\theta})
\end{equation}
This assumption is crucial when coupled with the prevalent iterative learning algorithms. In particular, it makes textbook machine learning analysis readily available \cite{Bishop2006MLBook}, where training, validation, and test data splits are compared to understand whether the model ``generalizes'' in evaluation. Unfortunately, such assessment of generalization is also limited to what is well represented and practically assessable in the split \cite{zhang2017understanding}. Similarly, popular algorithms such as variational inference in autoencoders \cite{kingma2014auto,kingma2019introduction} rely on mean-field theory, e.g., measuring divergences to the mean and standard deviation of a (Normal) distribution to learn about the data generating process. 

The rooting of averages in these machine learning foundations entails several problematic technical factors. First, relying on aggregate measures exacerbates overconfident (false) predictions. In fact, models themselves are typically trained to give maximal predictions (or minimum losses) and as such seldom give out anything that is far away from the observed average value (which in labeled scenarios is typically a 100\% confidence of a category) \cite{Hendrycks2017SoftmaxOOD,Ovadia2019TrustUncertainty} \footnote{The observation of misleading overconfident predictions and the entailed false sense of evaluation correctness has empirically been made for both discriminative and generative models \cite{Nalisnick2019DoDeepGenerativesKnow} and further for model types beyond neural networks, such as probabilistic circuits \cite{Ventola2023DoPCsKnow}.}. Second, relying too much on averaging in generative modeling can either ``smooth out'' the diversity represented in the data, e.g., we create an envelope around two distribution modes, or approximate a particular mode well at the expense of dropping another, i.e., mode collapse \cite{kristiadi2016kl}. Finally, averaged measures as targets for evaluations incentivize the conception of systems that leverage shortcuts. For instance, the popular example of CleverHans predictors \cite{Lapuschkin2019CleverHans} has shown that an average accuracy measure does not allow us to distinguish whether images of a horse are classified correctly because they indeed contain a horse, or because they contain a different, potentially unidentifiable common feature (here, a photographer's tag). These confounders are particularly problematic in non-i.i.d. scenarios, where certain groups of features can become over or underrepresented at specific points in time \cite{busch2024truthriskgettingconfounded}, relating back to our section two's Gemini example. Aggregate measures thus make machine learning seem straightforward to evaluate and allow us to compare models, but the fact that these are imbued in our fundamental technological stack also severely limits prospective assessment.  \\

\noindent \textbf{Recommendations ---}\textit{towards individualized socio-technical assessment:} \\
A central recommendation is to avoid drawing general conclusions from assessments relying on averaged metrics. Although it seems convenient to report a single metric and be able to compare it to other works in the literature, the assertion of the average as being correct legitimizes the agenda of the majority \cite{Hermens1958TyrannyMajority}. For AI to find diverse applications, it must shift power to include what is different from the average. From a technical perspective, this may include steering away from classically employed statistics \cite{VonMises1964Statistics}, to the analysis of individual samples or extreme values \cite{Boult2019LearningUnknown}, or to re-investigate tilted loss functions \cite{Li2023TiltedLosses}. The latter exist in theory but are seldom used. We additionally recommend the use of multi-dimensional evaluation strategies, e.g., in the form of ``compasses'' \cite{ghosh2024simp,Mundt2022ClevaCompass}. These tools allow inspection of several dimensions of interest and should continue to be developed for transparent socio-technical assessment.

\subsection{``(Over-)Selling'': the abundant surrogate impasse}
\begin{quote}
    \emph{''Machine learning models routinely achieve perfect performance in one dataset even when that dataset is a large international multisite clinical trial. However, when that exact model was tested in truly independent clinical trials, performance fell to chance levels. Even when building what should be a more robust model by aggregating across a group of similar multisite trials, subsequent predictive performance remained poor.``} \citet{Chekroud2024IllusoryGeneralizabilityMedical}, editor summary - Peter Stern.
\end{quote}

\noindent The above quote highlights strong concerns over practical AI system deployability, despite large-scale trials. Although the quote also hints at our earlier section's issue with aggregates, we posit that an additional technical aspect is overlaid. We term this challenge the ``abundant surrogate impasse'', pointing to a lack of understanding of how optimization goals relate to practical measurement. In turn, we posit that even the most rigorous approach is technologically challenged in its assessment of practical implications, contributing to the overselling of our AI systems beyond human intent.

From a technical perspective, almost any machine learning system needs to be optimized via a proxy. This is both motivated by the fact that we require smooth and differentiable loss functions to obtain learning signals \cite{Murphy2012ProbMLBook} (e.g. turning a categorical 0 or 1 signal in classification into a spectrum between 0-1), and the fundamental limitation that we can rarely express our goal directly mathematically. Take for instance two prominent recent advances, generative vision models and large language models. In the former, we wish to train a model that is capable of faithfully generating a diverse set of images, yet it is unclear how to express this goal directly. As such, it is frequent, for instance in auto-encoding based models, to minimize the discrepancy in pixel values of an original $\mathbf{x}$ and a reconstructed image $\mathbf{\hat{x}}$:  
\begin{equation}
    \mathcal{L}(\mathbf{\theta}) = {\textstyle\sum\nolimits_{n=1}^{N}} ||\mathbf{\hat{x}}_n - \mathbf{x}_n||_2^2 \quad \mathtt{where} \quad \mathbf{\hat{x}}_n = h_{\mathbf{\theta}}(\mathbf{x}_n)
\end{equation}
Similarly, in large language models, we wish to accurately model language, yet we don't explicitly encode linguistic rules or semantic coherence as training objectives. 
Instead, the standard approach involves tokenizing text into discrete units, treating each token from the vocabulary as a distinct class, and training the model to predict the next token ($\mathbf{x_{t+1}}$) in a sequence ($\mathbf{x_{t:1}}$) through maximum likelihood estimation. 
This seemingly simple objective of next-token prediction serves as a proxy for learning deeper linguistic patterns and relationships.
\begin{equation}
    \mathcal{L}(\mathbf{\theta}) = - {\textstyle\sum\nolimits_t} \log p_{\mathbf{\theta}}(\mathbf{x}_{t+1}|\mathbf{x}_{t:1})
\end{equation}

In both cases, these training objectives and loss functions are used to optimize the system and evaluate its performance on held-out test data. 
However, when assessing real-world generation capabilities, we typically sample new outputs and evaluate their quality. This assessment requires fundamentally different metrics - perceptual scores for images \cite{kazmierczak2022study}, reference-based \cite{papineni2002bleu, lin2004rouge} and reference-free metrics for languages \cite{zhang2020bertscoreevaluatingtextgeneration}.

Thus, it strikes us as unsurprising that AI systems are commonly oversold. Some real-world concepts remain unmeasurable or their complexity cannot be measured through a single value. A single loss proxy, where we are unable to express our true goal, is used for training, and a set of different measures is brought out in evaluation. In turn, the narrow optimization focus lacks relation to the actual world, but our often desired abstract concepts are hard, if not impossible, to conceptualize mathematically. The discrepancy between what is being optimized for, what is desired as the outcome, and what is being evaluated, can lead to questionable conjectures of a system's capabilities. \\

\noindent \textbf{Recommendations ---}\textit{towards meaningful assessment of outcome and impact:} \\
Our final central recommendation is to avoid generalized conclusions originating from evaluation of limited proxies. We recommend particular caution in modern systems where optimization objectives and evaluation measures are mismatched. In addition, we raise awareness for the fact that we no longer know if data contamination has already provided the system with answers during training \cite{Borji2023ChatGPTFailures}, or the reality that much of what is categorized should be considered fluid and complex (social) constructs, i.e., the employed proxy should not have been used from the start. 
Instead of drawing wide conclusions from measured proxy values, we recommend to focus attention on effective outcomes, for instance, akin to a ``social impact dashboard'' \cite{ghosh2024simp}. This in turn requires a shift away from learning and assessing correlations, to gaining understanding in the form of ``Were that factor different, would the outcome change?'' \cite{Pearl2009Causality}. Causality is but one field that attempts to provide answers for such counterfactual questions. However, more research is necessary to understand when conditions for causal analysis and interventions are realizable in practice \cite{Kusner2017CounterfactualFairness}.

\section{Limitations and Disclaimer}
\begin{quote}
	\emph{``Most people think the issue is changing social norms. It is, but it’s also ($\ldots$) how willing engineers are to change those systems.''} Meredith Broussard \cite{broussard2023more}
\end{quote}

\noindent Our work does not claim that social challenges will eventually be overcome through advances in AI design alone. On the contrary, resonating with above quote, we believe that making AI systems more socially and technologically sustainable, requires \textit{both} the willingness for people to change social norms and to reconceive technical tools to allow this change. 
With this in mind, there exist a plethora of societal and ethical aspects that our re-conceptualization of cake as a metaphorical AI system has either referenced only in passing or is unable to cover. This is because our work's focus is on pointing out prevalent ramifications and how they are limited by foundational technical underpinnings. As such our work serves the primary purpose to raise awareness of how social and technical elements are inevitably intertwined and, through mapping of social outcomes to technical foundations, create opportunity for different communities to engage in cross-disciplinary dialogue along actionable dimensions. However, we acknowledge that this approach, despite highlighting important new avenues, is also inevitably bound by the cake analogy as a frame of reference. In particular, we understand that the idea of baking a cake, whether reconceived or not, may not be what is desirable on many occasions. To re-emphasize our portions on (over-)selling, there is a crucial difference between naive over-claiming of capabilities and deliberately deploying AI in unnecessary or harmful places.
In a similar spirit, our work remains subject to a host of further factors, including, but not limited to, monetary incentives, pressure to publish, the challenges of the academic reviewing system, and imbalanced power dynamics. 

\section{Conclusion}
We have detailed how the process of making a cake serves as a comprehensive analogy to the design of modern AI systems. Through several drawn parallels across the full AI life cycle, we have highlighted why change towards more sustainable and collaborative AI systems is not merely a social challenge, but how it is also constrained by prevalent technical foundations. Ultimately, our analysis and recommendations thus call for a departure from the traditional ML textbook narrative \cite{Bishop2006MLBook,Murphy2012ProbMLBook,Goodfellow2016DLBook} to equally focus on process improvements and collective exploration \cite{Birhane2022powertopeople}. However, such participatory practice \cite{Arnstein1969ladder} that spans \emph{consultation, inclusion, collaboration} and \emph{ownership} seems to presently be difficult to implement in practice \cite{Delgado2023participatoryturn, AdaLoveInst2021ParticipatoryDataSteward} and must also be embraced technologically. We believe that our re-conceptualization of the AI cake analogy serves as an opportunity to engage in deeper cross-disciplinary dialogue towards this goal. As such, we envision future research to continue the analysis of the ties between technical limitations and social ramifications, eventually re-imagining the present process of baking our metaphorical AI cake. 
\clearpage

\bibliographystyle{ACM-Reference-Format}
\bibliography{aaai24}

\end{document}